\documentclass[11pt]{article}
\usepackage{coling2020}
\usepackage{times}
\usepackage[hyphenbreaks]{breakurl}
\usepackage[hyphens]{url}
\usepackage{latexsym}

\usepackage{tabularx}

\usepackage{mathptmx}
\usepackage[scaled=.90]{helvet}

\usepackage{makecell}
\usepackage[table]{xcolor}

\usepackage{multirow} 
\usepackage{amsmath}

\usepackage{amssymb}

\usepackage{float}

\usepackage[utf8]{inputenc}

\usepackage{listings}
\usepackage{graphicx} 
\usepackage{graphics}

\usepackage{textcomp}

\usepackage{tabstackengine}

\colingfinalcopy

\title{Social Media Unrest Prediction during the COVID-19 Pandemic:\\ Neural
Implicit Motive Pattern Recognition\\ as Psychometric Signs of Severe Crises}

\author{Dirk Johannßen \\
 MIN Faculty \\
 Dept. of Informatics \\
 Universität Hamburg \\
 \& Nordakademie \\
 \hspace{7cm}\url{http://lt.informatik.uni-hamburg.de/} \\
 \hspace{7cm}{\tt \{biemann, johannssen\}@informatik.uni-hamburg.de}\\
 \\\And
 Chris Biemann \\
 MIN Faculty \\
 Dept. of Informatics \\
 Universität Hamburg \\
 22527 Hamburg, Germany \\
}

\date{}

\begin{document}
\maketitle
\bigskip 

\begin{abstract}
 
 The COVID-19 pandemic has caused international social tension and unrest. 
 Besides the crisis itself, there are growing signs of
 rising conflict potential of societies around the world. Indicators of global
 mood changes are hard to detect and direct questionnaires suffer from social 
 desirability biases. However, so-called implicit methods can reveal
 humans intrinsic desires from e.g.~social media texts. We present
 psychologically validated social unrest predictors and replicate scalable and
 automated predictions, setting a
 new state of the art on a recent German shared task dataset. We employ this
 model to investigate a change of language towards social unrest during
 the COVID-19 pandemic by comparing established psychological
 predictors on samples of tweets from spring 2019 with spring 2020.
 The results show a significant
 increase of the conflict-indicating psychometrics. With this work, we
 demonstrate the applicability of automated NLP-based approaches 
 to quantitative psychological research.
\end{abstract}

\section{Introduction}\label{sec:intro}
\blfootnote{

\hspace{-0.65cm}  
This work is licensed under a Creative Commons
Attribution 4.0 International License.
License details:
\url{http://creativecommons.org/licenses/by/4.0/}.
}

The COVID-19 pandemic and the reactions to it have led to growing social
tensions.
Guitérrez-Romero~\shortcite{gutierrez-romero_conflict_2020} studied the effects
of social distancing and lockdowns on riots, violence against civilians, and
food-related conflicts in 24 African countries. The author found that the
risk of riots and violence have increased due to lockdowns. 
Resistance against national health regulations such as the duty to wear masks are partially
met with resistance by movements such as
anti-maskers or anti-obligation
demonstrations.\footnote{\url{https://firstdraftnews.org/latest/coronavirus-how-pro-mask-posts-boost-the-anti-mask-movement/}} 
Even anti-democratic alterations of e.g.~services offered by the US Postal
Service (USPS) of delivering mail-in ballots for the US presidential elections 2020, which are essential for social distancing measures
amidst the pandemic, are being utilized amidst this international crisis and
foster social unrest and potential outbursts of violence, civil disobedience or
uprisings.\footnote{\url{https://www.businessinsider.com/trump-walks-back-threat-block-covid-relief-over-usps-funding-2020-8?r=DE&IR=T}}

Social media has become an important reflection of nationally and
internationally discussed topics, and is a predictor of e.g.~stock markets,
disease outbreaks or political elections \cite{kalampokis_understanding_2013}.
The majority of human-produced data exists in textual form and broadly in social media and thus, an
the investigation of social unrest and conflict situations from social media
becomes a worthwhile application area for natural language processing (NLP)
problems \cite{gentzkow_text_2019}.

When speaking about such global phenomena such as a rise in international social
unrest and possible occurrences of conflict reflected in text, the detection of
specific keywords or utterances have not been successful in past research.
Mueller et al.~\shortcite{mueller_reading_2017} utilized Laten Dirichlet
Allocation (LDA, \cite{blei_latent_2003}) topic modelling on war-related
newspaper items and were not able to improve predictability from other
multi-factor models that take into account e.g. GDP figures, mountainous
terrain or ethnic polarization. Furthermore,
Chadefaux~\shortcite{chadefaux_early_2012} showed that news reports on possible
war situations alone did not function as good predictors but identified sharp
frequency increases before war emerged, possibly helping with just-in-time
safety measures but likely failing to avoid war situations altogether.

Alternatively, the risks of escalation could be determined based on
politician's personalities and the current mood and tone of utterances \cite[p.
407]{schultheiss_implicit_2010}. However, intrinsic desires and personality can
hardly be measured directly (see Section \ref{sec:implicit_motives}).
Intrinsic or subconscious desires and motivation would more likely correlate
with personalities, tone, and thus possibly social unrest.

We hypothesize that the frequency of social unrest predictors have significantly
changed in social media textual data during the COVID-19 pandemic drawn from the Twitter 1
percent stream\footnote{\url{
https://developer.twitter.com/en/docs/labs/sampled-stream/overview}} in 
early 2019 and 2020, whilst linguistic features stay comparably stable and unchanged. With
this, we aim to demonstrate a possible transition from
laborsome manual psychological research to automated NLP approaches.

After presenting and discussing related work in Section \ref{sec:related_work},
we will first introduce the concept of implicit motives and self-regulating
levels in more details in Section \ref{sec:implicit_motives} and the social
unrest predictors thereafter in Section \ref{sec:social_unrest_predictors}. The data utilized for experiments
is described in Section \ref{sec:data} and the methodology 
in Section \ref{sec:methodology}. Thereafter, we
will present the results in Section \ref{sec:results} and discuss their impacts in Section
\ref{sec:discussion}. Lastly, we will draw a conclusion in Section
\ref{sec:conclusion}.

\section{Related Work}\label{sec:related_work}

Conflict predictions from natural language are rarely encountered
applications and have mainly been about content analysis and less about crowd psychology.
Kutuzov et al.~\shortcite{kutuzov_one--x_2019} used one-to-X analogy reasoning based on word embeddings for predicting previous
armed conflict situations from printed news. Johansson et
al.~\shortcite{johansson_detecting_2011} performed named entity recognition
(NER) and extracted events via Hidden Markov Models (HMM) and neural networks,
which were combined with human intelligence reports 
to identify current global areas of conflicts, that, in turn, were utilized
mainly for world map visualizations.

Investigation of personality traits has
mainly been focussing on so-called explicit methods. For these, 
questionnaires are filled out either by
interviewers, through observations, or directly by participants. One of the most
broadly utilized psychometrics is the Big Five inventory, even though its
validity is controversial \cite{block_contrarian_1995}.
The five-factory theory of personality (later named Big Five) identifies five personality traits,
namely \emph{openness to experiences}, \emph{conscientiousness},
\emph{extraversion}, \emph{agreeableness} and \emph{neuroticism}
\cite{mccrae_five-factor_1999,goldberg_language_1981}. This Big Five inventory
was utilized by Tighe and Chegn~\shortcite{tighe_modeling_2018} for analyzing
these five traits of Filipino speakers. 

Some studies perform natural
language processing (NLP) for investigating personality traits. Lynn et
al.~\shortcite{lynn_hierarchical_2020} utilized an attention 
mechanism for deciding upon important parts of an instance when 
assigning the five-factor inventory classes. The Myers-Briggs 
Type Indicator (MBTI) is a broadly utilized
adaption of the Big Five inventory, which
Yamada et al.~\shortcite{yamada_incorporating_2019} employed for asserting the
personality traits within tweets.\footnote{A \emph{Tweet} is a short message from
the social network microblogging service Twitter (\url{https://www.twitter.com/})
and consists of up to 240 characters.}

The research field of psychology has moved further towards automated language
assertions during the past years. One standard methodology is the utilization of
the tool linguistic inquiry and word count (LIWC), developed by Pennebaker et
al.~\shortcite{pennebaker_linguistic_1999}. The German version of LIWC was
developed by Wolf et al.~\shortcite{wolf_computergestutzte_2008}. It includes 96 target classes,
some of which are rather simple linguistic features (word count, words longer
than six characters, frequency of punctuation), and psychological categories such
as anxiety, familiarity, or occupation. Even though the tool appears rather
simple from an NLP point of view, it has a long tradition to be utilized for
content research in the field of behavioral psychology. Studies
utilizing LIWC have shown that function words are valid predictors for long-term 
developments such as academic success \cite{pennebaker_when_2014}.
Furthermore, it has been shown that LIWC correlates with the Big Five inventory
\cite{mccrae_five-factor_1999}. Importantly, the writing style of 
people can be considered a trait, as it has shown high stability over time, which means that
it is not dependent on one's current mood, the time of day, or other external
conditions \cite{pennebaker_linguistic_2000}.
Hogenraad~\shortcite{hogenraad_words_2003} utilized an implicit motive
(see Section \ref{sec:implicit_motives}) dictionary approach to automatically 
determine risks of war outbreaks from different novels and historic documents,
identifying widening gaps between the so-called power motive and affiliation
motive in near-war situations.

Overall, the work on automated classification of implicit
motive data or the use of NLP for the assertion of psychological traits in
general is rather sparse or relies on rather outdated methods, as this
application domain can be considered a niche
\cite{schultheiss_implicit_2010,johannsen_neural_2019,johannsen_between_2018,johannsen_reviving_2019}.
One recent event in this area was the GermEval 2020 Task 1 on the 
Classification and Regression of Cognitive and Motivational Style from Text \cite{johannsen_germeval_2020}. 
The best participating team reached a macro f1 score of 70.40 on
the task of classifying implicit motives combined with self-regulating levels, 
resulting in 30 target classes. However, behavioral outcomes from automatically
classified implicit motives have -- to our knowledge -- not yet been researched.

\section{Implicit Motives and Self-regulatory
Levels}\label{sec:implicit_motives}

\begin{figure}[!ht]
	\begin{center}
		\includegraphics[width=0.25\textwidth]{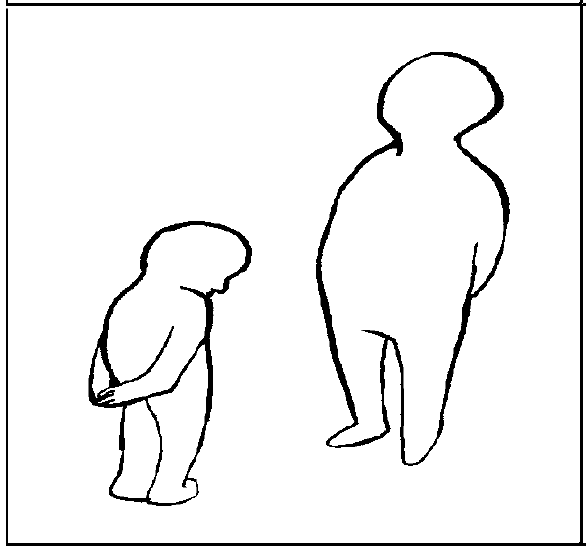}
		\caption{Exemplary image to be interpreted by participants utilized
		for the operant motive test (OMT). Identifiable motives are the
		affiliation motive (A), the power motive (M), achievement (L), and freedom
		(F). A 0 represents the unassigned motive \cite{kuhl_operante_1999}.}
		\label{fig:omt}
	\end{center} 
\end{figure}

Implicit motives can reveal intrinsic,
unconscious human desires, and thus avoid social desirability biases, which
usually are present when utilizing direct questionnaires.
They originated from the Thematic Apperception Test (TAT) by Murry et al~\shortcite{murray_thematic_1943}.
Participants are confronted with ambiguous images of multiple people that interact with each 
other as displayed in Figure \ref{fig:omt}, and are asked to answer four
questions: i) who is the main person, ii) what does that person feel? iii) 
why does the person feel that way, and iv) how
does the story end? From these questions, trained psychologists can assign one
of five motives: affiliation (A), freedom (F), achievement (L), power (M), and
zero (0). The psychologists follow some rules, one being the so-called
primacy rule, where the very first identifiable motive determines the whole
instance, despite what follows \cite{scheffer_auswertungsmanual_2013}. These
motives have shown to be behavioral predictors and allow for 
long-term statements of e.g.~group dynamics or success
\cite{mcclelland_leadership_1982,schultheiss_implicit_2010}. 
Implicit motives have been broadly utilized in the 1980s but at the cost of 
laborsome manual annotating processes. It takes about 20 hours of training for an
annotator to encode one of the implicit motives. Skilled human annotators 
take up to 50 hours per 100 participants. This costly
annotating process has hampered this once-promising psychometric \cite[p.
140]{schultheiss_implicit_2010}. 

Whilst the classification performance of implicit motive models have been
explored and achieved high results (e.g.~\cite{johannsen_neural_2019,johannsen_germeval_2020}), behavioral
consequences and mass phenomena from automated labeled textual instances have barely been researched.

In addition to the implicit motives, the data set from the GermEval20
Task 1 comes with so-called levels per textual instance. The levels were
developed by Kuhl~\shortcite{kuhl_motivation_2001}. They describe the
self-regulatory enactment in five dimensions. According to Scheffer and
Kuhl~\shortcite{scheffer_auswertungsmanual_2013} the 1st level is the 
ability to self-regulate a positive
affect, the 2nd is the sensitivity for positive incentives, the 3rd 
self-regulates a negative affect, the 4th is the
sensitivity for negative incentives and the 5th level describes the passive
coping with fears. In other words, these levels help to identify
the type of the participant's emotional response 
according to the identified implicit motive. 

As with many psychometrics, the reliability of implicit motives, and
especially their predecessor (the TAT) is
controversial. One main point of criticism is that implicit motives do not
correlate significantly with so-called explicit motives. Whilst implicit motives
try to measure intrinsic desires indirectly by asking participants associative
questions, explicit motives try to measure desires via direct questionnaires. In
psychology, reliability means, that personality traits revealed by one measure
may not conflict with personality traits measured by other, well-established
measures. Since the measured desires of implicit and explicit motives generally
do not match, the reliability of implicit motives is said to be weak.

Schultheiss et al.~\shortcite{schultheiss_implicit_2010} explain this lack of 
reliability and correlation with the fact that explicit implicit motives are by 
definition of different measurements that can not be directly compared. Whilst
implicit motives measure intrinsic desires that are subconscious, explicit
motives are more influenced by a social expectation bias (i.e.~what do
participants think is a socially sound and accepted answer to a question) and
thus are closer connected to behaviorism. Nonetheless, reliability in
psychological research demands different observable results of metrics 
to be coherent \cite{reuman_ipsative_1982} when the descriptions of what a
psychometric is supposed to measure matches (i.e.~desires).

Another point of criticism is the way the TAT images are selected. They emerge 
from an empirical study, where participants are shown different images. Only 
when past frequencies of motives are achieved with an image, this image gets 
added to the available
testing stock. With this, however, the very first selected implicit motive
images could not have been validated. Nowadays, many scholars argue that the
amounts of positive evidence legitimize this methodology, but it has yet to be
resolved \cite{hibbard_critique_2003}.

The transition from natural language to
intrinsic desires and motivation is not trivial, as humans do not 
express intrinsic and unconscious desires unfiltered and directly. As soon as a direct
questionnaire is involved, social desirability biases (i.e.
thoughts of publicly expected answers) alter an uninfluenced introspection
\cite{brunstein_implicit_2008}. Such direct questionnaires are called
explicit methods, in contrast to implicit methods, such as e.g.~the TAT and
subsequent tests produced by image descriptions.

\section{Social Unrest Predictors}\label{sec:social_unrest_predictors}

Times of severe social unrest are reflected by distinct patterns of implicit
motives and linguistic features. Winter~\shortcite{winter_role_2007}
surveyed multiple prior studies, identifying three main predictors:
\emph{responsibility}, \emph{activity inhibition}, and \emph{integrative
complexity}, displayed in Table \ref{tab:social_unrest_metrics}. 
In this study, the author
identified and analyzed 8 occurrences of crises and social unrest by examining
influential political speeches of this time. Thereafter, the outcomes of these crises -- whether they ended peacefully
or in conflict -- were projected on indicators from earlier research.

Winter and Barenbaum~\shortcite{winter_responsibility_1985} found that the 
power motive (M) has a moderating effect of
responsibility. In
other words, responsibility determines, how vast amounts of power motivated
expressions are behaviorally enacted. If a high responsibility score
is measurable, power motivated individuals act pro-social. On the contrary, if
the responsibility score is low, aggression and lack of leadership are to be
expected. 

\begin{table}[!h] 

\begin{tabular}{|p{4.5cm}|p{5cm}|p{5cm}|}\hline
 \textbf{Category} &
 \textbf{Measure} &
 \textbf{Example or Explaination} \\\hline
\multicolumn{3}{l}{ } \\
\multicolumn{3}{c}{Responsibility} \\\hline
 i) moral standards &
 observable, if people, actions, or things are described with either morality or legality &
 'she wants to do the right thing' \\\hline
 
 ii) obligation &
 means, that a character in a story is obliged to act because of a rule or regulation &
 'he broke a rule' \\\hline
 
 iii) concern for others &
 emerges, when a character helps or intends to help others or when sympathy is
 shown or thought & 'the boss will understand the problem and will give the worker a raise'
 \\\hline
 
 iv) concerns about consequences &
 can be identified when a character is anxious or reflects upon negative outcomes &
 'the captain is hesitant to let the man on board, because of his instructions'
 \\\hline
 
 v) self-judgment &
 scores when a character critically judges his or her value, morals, wisdom, self-control, etc. and has to be intrinsic &
 'the young man realizes he has done wrong' \\\hline
\multicolumn{3}{l}{ } \\
\multicolumn{3}{c}{Activity inhibition} \\\hline
 linguistic negation &
 in English terms, the authors describe activity inhibition as the frequency of
 ``not'' and ``-n't'' & responsibility measure, e.g.~a variable negatively correlated with male
 alcohol consumption \\\hline
 
 leadership motive pattern (LMP) &
 combined with a high power motive (M) and low affiliation motive (A) &
 predicts responsible leadership power behaviors instead of profligate
 impulsive expressions of power \\\hline
\multicolumn{3}{l}{ } \\
\multicolumn{3}{c}{Integrative complexity} \\\hline
\multirow{2}{*}{\begin{tabular}[c]{@{}l@{}}7-point
continuum\\ range score from\\ simplicity to\\ differentiation\\ and
integration\end{tabular}} & 1: no sign of conceptual and differentiation or
integration can be observed & only one solution is considered to be legitimate
\\\hline 
 &
 7: overreaching &
 viewpoints are expressed, involving different relationships between alternate
 perspectives\\\hline
\end{tabular}
\caption{According to Winter ~\shortcite{winter_role_2007}, some distinct
psychometrics and their combinations predict social unrest -- namely
responsibility, activity inhibition, and integrative complexity. The table shows
their categories, measurements and offer examples or explanations.
Responsibility is measured with a dedicated TAT, activity inhibition (AI) and
integrative complexity is determined via content analysis. Especially the 
combination of low responsibility, high activity 
inhibition and little integrative complexity (e.g.~
high frequency of the power motive combined with the self-regulatory 4th level)
predict situations of social unrest with negative escalatory outcomes.}
\label{tab:social_unrest_metrics}
\end{table}

\emph{Activity inhibition} is reflected, according to by McClelland et
al.~\shortcite{mcclelland_drinking_1972} as the frequency of ``not'' and
``-n't'' contradictions in TAT or other verbal texts. Activity inhibition
functions as motivational and emotional regulation. The authors identified a
negative correlation between activity inhibition and male alcohol consumption.
Combined with a high power
motive (M) and low affiliation motive (A), subsequent research by McClelland and
his colleagues revealed a so-called leadership motive pattern (LMP)
\cite{mcclelland_leadership_1982,mcclelland_human_1988}.
The higher this LMP, the more responsible leaders act. As for \emph{integrative
complexity} it was observed, that the lower the frequency of utterances
in accordance to the 7-point score was (see Table
\ref{tab:social_unrest_metrics}), the more likely escalations became.
 
\section{Training and experimental data}\label{sec:data}

For testing the proposed hypothesis (Section \ref{sec:intro}), we first train a
classification model and utilize this model for testing social network textual
data. In this section, we will describe the two different data sources for
training and the experiments.
 
\subsection{Model Training Data}
The data utilized for training models were made available by the organizers of
the \emph{GermEval-2020 Task 1 on the Classification and Regression of
Cognitive and Motivational Style from
Text}\footnote{GermEval is a series of shared task evaluation campaigns
that focus on Natural Language Processing for the German
language.}\footnote{\url{
https://www.inf.uni-hamburg.de/en/inst/ab/lt/resources/data/germeval-2020-cognitive-motive.html}}
\cite{johannsen_germeval_2020}.
The training set consists of 167,200 unique answers, given by 14,600
participants of the OMT (see Section \ref{sec:implicit_motives}). The training
data set is imbalanced. The power motive (M) is the most frequent class,
covering 41.02\% of data points.
The second most frequent class, achievement (L) only accounts for 19.63\% and thus is half as frequent as M. The training data was assembled and
annotated by the University of Trier, reaching a pairwise annotator intraclass
correlation of r = .85. With only 22 words on average per training instance
(i.e.~a participant's answer) and a standard deviation of 12 words, training a
classifier on this data is a short text classification task \cite{johannsen_germeval_2020}.\footnote{The data can be retrieved via
\url{https://www.inf.uni-hamburg.de/en/inst/ab/lt/resources/data/germeval-2020-cognitive-motive.html}}

\subsection{Experimental Data}\label{subsec:experimental_data}

The experimental data was collected before this work by crawling the Twitter
API and fetching 1 percent of the worldwide traffic of this social network
\cite{gerlitz_mining_2013}. We sample posts over the time window from
March to May of both, 2019 and 2020. There are no apparent linguistic
differences between the two samples. The average word count, part-of-speech
(POS) tags, sentence length, etc. are comparable. 

Thereafter, we extracted the \emph{text} and \emph{date time} 
fields of posts marked as German. From those files hashtags, name 
references (starting with '@'), corrupted lines, and any post shorter than three
content words were removed. The resulting files 
for 2019 and 2020 contained more than 1 million instances. Lastly, the instances
were randomly shuffled. We drew and persisted 5,000 instances per year for
the experiments, as this data set size is large enough
for producing statistically significant results.
The posts on average consist of 11.97 (2019) and 11.8 (2020) words per sentence,
and thus are very short. During the experiments, further pre-processing steps
were undertaken, which are described in Section \ref{sec:methodology}. By
stretching out the data collection time window and by comparing the same 
periods in two subsequent years, we aim to reduce any bias effect that might impact Twitter user behavior over short periods, e.g.~
the weather, any sports event, or short-lived political affairs.

\section{Methodology for Implicit Motive Classification Social Unrest
Prediction}\label{sec:methodology}

For constructing a model of sufficient quality to test our hypothesis, we
follow Johannßen and Biemann~\shortcite{johannsen_neural_2019} and train a long
short-term memory network (LSTM, \cite{hochreiter_long_1997}) 
combined with an attention mechanism. 

An LSTM is a special type of recurrent
neural network (RNN). An RNN not only has connections between
units from layer to layer but also between units of the same layer. Furthermore an
LSTM also has a mechanism called the \emph{forget gate}, allowing the structure
to determine which information to keep and which information to forget during
the training process.
The attention mechanism \cite{young_recent_2018} can capture the intermediate importance of algorithmic decisions made by the
network. It can be employed for enhanced results but also investigated for
researching algorithmic decisions. However, it is debated upon, whether this
algorithmic importance can serve as an explanation. Even though oftentimes, the
algorithmic importance is correlated with an explanation for the task
(i.e.~does a model for image recognition of animals \emph{look} at the animals
or the backgrounds of the images?), there are
cases, where algorithmic
importance and explanation for the task
differ \cite{jain_attention_2019,wiegreffe_attention_2019}).
Since automatically labeling implicit motives is a sequential problem
revolving around identifying the first verbal enactment of a motive (see Section
\ref{sec:implicit_motives}, we decided to employ a Bi-LSTM with an attention mechanism \cite{schuster_bidirectional_1997}.

We decided against additional features
such as part of speech (POS) tags or LIWC features like in our previous work
\cite{johannsen_neural_2019}, as we did not reach the best results with these
additional features. The maximum token length was set to 20, as determined by
preliminary experiments \cite{johannsen_neural_2019}, and
reflects the primacy rule of the implicit motive theory described 
in Section \ref{sec:implicit_motives}. The average answer length of
the training data set was 22 tokens (see Section \ref{sec:data}). With this
decision to limit the considered tokens, we aim to closely replicate the
implicit motive coding practices manually performed by trained psychologists
\cite{kuhl_operante_1999}. Accordingly, it is preferable to assign the 0 motive
(i.e.~no clear motive could be identified) than to falsely assign a motive that
is not the very first one in the sequence.

Some standard pre-processing steps were applied to reduce noise, which
was to remove the Natural Language Toolkit (NLTK) German corpus stop words\footnote{The Natural Language Toolkit (NLTK) is a collection of python libraries for NLP
\url{https://www.nltk.org/}.}, to lowercase the text, remove numbers, normalize
special German letters (i.e.~umlaute). Emojis were removed as well, since
Twitter offers a selection of
3,348 emojis\footnote{\url{https://emojipedia.org/twitter/twemoji-13.0.1/}} ,
that in turn mainly do not capture sufficient informational gain per textual 
answer for the task at hand.
To remove stop words has to be an informed choice when it comes to 
performing NLP on psychological textual data. For example, function words are
said to predict academic success (see Section \ref{sec:related_work}). However, during our
experiments, we saw an increase in model performance when stop words were
removed.

After the training, we utilize the model on the two sampled data
sets described in Subsection \ref{subsec:experimental_data}. According to our
hypothesis in Section \ref{sec:intro} and following the theories in Section
\ref{sec:social_unrest_predictors}, we investigate the frequency of the power
motive with the self-regulatory level 4, which we expect to be higher. At the
same time, we will also analyze the other motives and levels to see which ones
are now less frequent and to what extent. Furthermore, we compare different
linguistic features and statistics from 2019 to 2020 to see, if any of these
show differences that might indicate possible biases in the data.

Our Bi-LSTM model was set to
be trained within 3 epochs and with a batch size of 32 instances. The model was
constructed having 3 hidden layers and utilized pre-trained fasttext embeddings
\cite{bojanowski_enriching_2017}, as this character-based or word fragment-based language representation
has shown to be less prone to noisy data and words that have not been observed yet like e.g.~spelling mistakes or slang -- both often
observable in social media data.
The fasttext embeddings had 300 dimensions and were trained on a Twitter corpus, 
ideally matching the task at hand.\footnote{The fasttext model was obtained from
Spinningbytes at \url{http://spinningbytes.ch/resources/wordembeddings}}
Explorative experiments with different parameter combinations have shown 
that a drop-out rate of .3 and step width of
.001 produced good results.

The cross-entropy loss was reduced rather quickly and oscillated at 1.1 when we
stopped training early during the second epoch. After each epoch, the model was
evaluated on a separate development test. After the training was finished, the model was
tested once on the GermEval20 Task 1 test data and with the official evaluation
script.
This provides the chance to compare the achieved results with the best-participating team. Schütze et al.~\shortcite{schafer_predicting_2020} achieved 
a macro f1 score of 70.40, which our Bi-LSTM model was able to
outperform with an f1 score of 74.08, setting a new state of the art on this
dataset.

\section{Results}\label{sec:results}

After having trained the Bi-LSTM model and sampled the experimental data, we
will describe the results and findings of the conducted Twitter COVID-19 experiments
in this section. An overview of all results is displayed in Table
\ref{tab:results}.

\begin{table}[!h]

\centering
\begin{tabular}{|r|r|r|r|r|}
\hline
\textbf{Metric} & \textbf{2019} & \textbf{2020} & \textbf{Percentage delta} &
\textbf{Significance} \\ \hline \multicolumn{5}{l}{ } \\
\multicolumn{5}{c}{Activity inhibition and responsibility} \\\hline
Power 4 & 33.76 & 37.40 & 10.97 & p\textless{}.01*** \\
\hline 
LIWC Family & .08 & .05 & -37.60 & p\textless{}.05* \\
\hline 
LIWC insight & .23 & .17 & -26.09 & p\textless{}.05*  
\\\hline 
\multicolumn{5}{l}{ } \\
\multicolumn{5}{c}{Implicit motives} \\\hline
Power motive & 65.84 & 68.24 & 3.64 & p\textless{}.01*** \\ \hline
Freedom motive & 20.28 & 17.72 & -12.63 & p\textless{}.01*** \\ \hline
Achievement motive & 6.80 & 7.00 & 2.94 & p\textgreater{}.05  
\\\hline 
Affiliation motive & 2.00 & 1.86 & -7.00 &
p\textgreater{}.05 \\\hline 
Null motive & 5.10 & 5.10 & .00 &
p\textgreater{}.05 
\\ \hline
 
\multicolumn{5}{l}{ } \\
\multicolumn{5}{c}{Self-regulatory levels} \\\hline 
Level 1 & 6.50 & 6.01 & -7.54 &
p\textgreater{}.05 \\\hline 
Level 2 & 2.76 & 3.26 & 18.12 &
p\textgreater{}.05 \\\hline 
Level 3 & 27.20 & 25.58 & -5.96 &
p\textless{}.01*** \\\hline 
Level 4 & 42.78 & 45.20 & 5.67 &
p\textless{}.01*** \\\hline 
Level 5 & 15.86 & 14.92 & -5.93 &
p\textless{}.05* \\\hline 

\multicolumn{5}{l}{ } \\
\multicolumn{5}{c}{Linguistic statistics} \\\hline 
Average words & 11.97 & 11.80 & -1.42 &
p\textgreater{}.05 \\ \hline Verbs & 1.19 & 1.22 &
2.52 & p\textgreater{}.05 \\ \hline Adjectives & .43  
& .43 & .00 & p\textgreater{}.05 \\ \hline Words
\textgreater 6 letters & 38.65 & 38.86 & .54 & p\textgreater{}.05 \\ \hline
\end{tabular}
\caption{Overview of the different psychometric and statistical results. *
represents significant results, *** represents highly significant results. All
combinations of motives and levels have been examined. Note that most motives,
levels, and statistical values stay constant. However, power 4 is more frequent, whilst
the freedom motive is less. As the linguistic statistic metrics stay
relatively stable, this indicates no observable sampling bias.}
\label{tab:results}
\end{table} 
 
To investigate the main predictor for social
unrest \emph{activity
inhibition} (see Section \ref{sec:social_unrest_predictors}), the power motive
(M) in combination with level 4 was counted. The self-regulatory 
level 4 describes the sensitivity for negative 
incentives (see Section \ref{sec:implicit_motives}). 
These measures are collected for all four data sets. Our Bi-LSTM model assigned
power 4 in 33.76\% of all cases for the Twitter sample from March to May of
2019, making this the most frequent label. However, for the data sample from 2020, power 4 is as frequent as 37.4\%, making
this an increase of 10.97\%. For calculating the significance of this rise, we
perform a t-test on the label confidences for the power motive with
self-regulatory level 4 for both, 2019 and 2020 with the 5,000 samples from each
year (see Section \ref{sec:data}).

The two-sample t-test on the confidence levels shows, that the rise
in frequency is statistically significant ($p <
0.05$ with $\bar{x}_{1}
= .27$, $\bar{x}_2 = .29$, $\sigma_{1} = .28$, $\sigma_{2} = .28$, $N_{1} =
5,000$ and $N_{2} = 5,000$).

The affiliation motive (A) is barely classified, covering only 2\% (2019) and
1.89\% (2020) of all instances. The slight decrease is not statistically significant
($p > .05$). The frequency of self-regulatory level 4 is elevated by 6.7\%. The
whole of all assigned power motive labels has only risen by 3.64\%, both having
risen less than the combination of the power motive and level 4 combined. The
strongest decline in frequency can be measured for the freedom motive with
-12.63\%. The other motives of affiliation, achievement, and null have barely
changed in comparison to 2019 with 2020. The same holds for the average
amounts of words per sentence, verbs, adjectives, and words containing at least 6
letters, all of which have barely changed, not indicating sampling biases. An
overview of the class frequencies is provided in Table
\ref{tab:class_frequencies}.
 
\begin{table}[!h]
\small
\centering
\begin{tabular}{|r|r|r|r|}
\hline
\textbf{Implicit motive} & \textbf{Frequency} & \textbf{Self-regulatory
level} & \textbf{Frequency} \\ \hline
\multicolumn{4}{l}{ }\\
\multicolumn{4}{c}{2019} \\\hline 
Power & 3,251 & 1 & 492 \\\hline
Affiliation & 141 & 2 & 193 \\\hline
Achievement & 414 & 3 & 1,487 \\\hline
Freedom & 9622 & 4 & 1,872 \\\hline
Zero & 232 & 5 & 724 \\\hline
 & & 0 & 232 \\\hline
 
\multicolumn{4}{l}{ }\\
\multicolumn{4}{c}{2020} \\\hline 
Power & 3,433 & 1 & 316 \\\hline
Affiliation & 90 & 2 & 151 \\\hline
Achievement & 203 & 3 & 1,259 \\\hline
Freedom & 923 & 4 & 2,233 \\\hline
Zero & 761 & 5 & 780 \\\hline
 & & 0 & 261 \\\hline

\end{tabular}
\caption{Overview of the class frequencies.}
\label{tab:class_frequencies}
\end{table} 

Since both, responsibility and integrative complexity can only be measured by
employing a specific TAT and a questionnaire, which would have to be performed
with each Twitter user, we can only investigate activity inhibition as a
combination of the power motive with the self-regulatory level 4. However, we
will review some psychological LIWC categories, that follow a close description
as the five categories of Winter's responsibility scoring system
\cite{winter_responsibility_1985}. Relevant LIWC categories for the 
responsibility is the combination of
\emph{family}, which are terms connected to expressions like 'son' or 'brother',
and \emph{insight}, which contain expressions such as 'think' or 'know',
representing self-aware introspection.
Family shows a significant decrease from 2019 (0.08) to 2020 (0.05) of -37.5\%.
The frequency of insight terms fell from 2019 (.23) to 2020 (.17) 
by -26\%, all of which are statistically significant changes ($p<0.05$ for
both categories).

\section{Discussion}\label{sec:discussion} 

We hypothesized that the social unrest predictors
by Winter~\shortcite{winter_role_2007}, namely \emph{activity inhibition},
\emph{responsibility}, and \emph{integrative complexity} are automatable and
reveal changes in natural language and signs of social unrest 
observable through the use of social media textual data connected to 
the COVID-19 pandemic.

The main research objective of this work is to find novel approaches to
automatically provide the community with red flags for growing tensions and
signs of social unrest via social media textual data. For this, \emph{activity
inhibition} is the main predictor. It consists of a distinct shift in implicit
motives. It is present when the frequency of the power motive with the
self-regulating level 4 (sensitivity for negative incentives, see Section
\ref{sec:implicit_motives}) is elevated and the affiliation motive is suppressed -- 
even though Winter~\shortcite{winter_role_2007} did not find clear evidence of
the latter. The comparable rise by 10.97\% ($p<0.01$) is an indicator of the social
tension of COVID-19 related social media posts.

Since other linguistic statistics, such as the average amounts of adjectives,
verbs, words per sentence, or words containing at least 6 letters have barely
changed, this indicates that the measurable differences in social unrest
predictors are content-based and not due to linguistic biases.

It is remarkable, that whilst the power motive has been labeled more
frequently, the frequency of the labeled freedom motive has declined by -12.63\%
from 2019 to 2020.
This freedom motive has barely been researched yet but has a close connection to the power
motive. Whilst power-motivated individuals desire control over their fellow humans
and their direct surrounding for the sake of control, freedom-motivated
individuals seek to express themselves and want to avoid any restraining
factors. Motives are said to be rather stable but can change over time
\cite{schultheiss_implicit_2010}. This change in motive direction could indicate
a roughening of verbal textual content and interpersonal communication. Example
utterances classified as freedom and power from 2019 compared with 2020 are
displayed in Table \ref{tab:motives_comparison_2019_2020}. 
 
The change of responsibility, as reflected in LIWC
categories, retreated by roughly 30\% from 2019 to 2020. This responsibility
indicates a personal involvement in topics and decisions, that we feel are
relevant for our surroundings. If this involvement diminishes, our interest in
participating in constructive solutions to problems does as well.
 
\begin{table}[!h]

\centering
\begin{tabular}{r r}
\begin{lstlisting}[language={[LaTeX]TeX}, label=lst:motives2019]
M 'RT @FrauLavendel: ist es wahr 
 dass schulleitungen den 
 schüler*innen drohen 
F RT @UteWeber: Nach einem relativ 
 unfeierlichen, 
 regionalen Offline-Tag aufs Sofa 
 sinken, wie von der Tarantel 
 gestochen aufspringen und zur...
A Weltbestseller "P.S. Ich liebe
 dich" bekommt einen zweiten Teil 
 https://t.co/9Ifl5CrNAP
 ------- Translation -----
M 'RT @FrauLavendel: is it true that
 principals threatens students
F RT @UteWeber: after a relatively
 un-celebrational, regional offline
 day, as bitten by a tarantula
 jumping up
A world best-selling book "P.S. Ich 
 liebe dich" gets a second part
 https://t.co/9Ifl5CrNAP
\end{lstlisting}
&
\begin{lstlisting}[language={[LaTeX]TeX}, label=lst:motives2020]
M Corona-Regeln im Saarland sind zum 
 Teil absurd und unverhältnismäßig 
F RT @kattascha: In den USA bekommen 
 viele Menschen keine Lohnfortzahlung 
 im Krankheitsfall. Das bedeutet: 
 Selbst bei Verdacht auf #COVID19 w...
A Wer einen Discord-Server sucht, 
 um entspannt mit seinen Kollegen 
 zu zocken oder gemeinsam abzuhängen 
 ist hier genau...
------- Translation -----
M the Corona rules for the Saarland
 are partially absurd and dis-
 proportionate
F RT @kattascha: in the US a lot
 of people don't receive continued
 pay in case of illness. That means:
 even in case of suspected #COVID19
A Whoever is looking for a Discord
 server for enjoyably game with
 their colleagues or chill together, 
 is in the right place...
\end{lstlisting} 
\end{tabular}
\caption{Some example tweets from 2019 (left) compared with 2020 (right). Whilst
the power motive (M) is more frequent in 2020, the freedom motive (F) became less
frequent. The affiliation motive (A) was very infrequent in both, 2019 and
2020. This signature indicates increased social unrest in 2020 in comparison
with 2019.}
\label{tab:motives_comparison_2019_2020}
\end{table}
 
\section{Conclusion and Outlook}\label{sec:conclusion}

With this work, we conducted a first attempt at automating psychometrics for
investigating social unrest in social media textual data.
The Bi-LSTM model combined with an attention mechanism of this work achieved an
f1 score of 74.08 on 30 target classes, making it state of the art on a
respective recent shared task dataset. With this model, we measured a
statistically significant rise in the power motive with self-regulating level 4, which reflects the social unrest predictor of
\emph{activity inhibition} in the direct comparison of the samples from March to
May of 2019 vs. 2020.

Furthermore, we investigated \emph{responsibility}, which shows significant
reductions during the COVID-19 pandemic, hinting at negative outcomes of
interpersonal and verbal communication on the social media platform Twitter.

This first approach most likely does not qualify for a real-world social
prediction system. Predictions of such a system can not yet be reliable enough
for deriving necessary actions from them. On the upside, implicit motives do not
only qualify for examining general socio-economic tensions, but can be applied
on an individual or small group scale. As an example, detecting tensions within
a small group can help to shape the group and guiding it into a better fit.
Furthermore, we advocate for combining implicit motives with sufficiently many
complementary psychometrics and content-based analysis e.g.~sentiment analysis,
topic modeling, or emotion detection.

Besides those combinations with other information sources for future work,
different sampling approaches and larger data set sizes should be utilized for 
reproducing findings and research
correlations with other social unrest predictors and indicators. In
this work, we have made the first steps towards understanding the
automation of psychological findings. Since only 5,000 samples were
drawn from a single social network platform, we advocate for
broadening this approach to include many more samples from wider
time windows paired with mixing the data sources. In addition to
that, deeper investigations into the linguistic variances between
times of so-called social unrest and more peaceful times should be
performed, as those could reveal patterns and characteristics of
time-specific utterances. 

Even though this work is only introductory, the observed correlations and
social unrest patterns are in line with an intuitive assumption of how language in 
social media data changes amid a pandemic.
Future work arises in the application of this methodology on other events and
crises, eventually providing a quantitative basis for implicit motive research.

\bibliographystyle{coling}
\bibliography{coling2020}

\end{document}